%% file: ijcai26.tex
\documentclass{article}
\usepackage{ijcai26}

\usepackage{times}
\usepackage{soul}
\usepackage{url}
\usepackage[hidelinks]{hyperref}
\usepackage[utf8]{inputenc}
\usepackage[small]{caption}
\usepackage{graphicx}
\usepackage{amsmath}
\usepackage{amsthm}
\usepackage{booktabs}
\usepackage{algorithm}
\usepackage{algorithmic}
\usepackage[switch]{lineno}

\usepackage{pifont}
\usepackage{multirow}
\usepackage{makecell}
\usepackage{colortbl}
\usepackage{arydshln}

\title{Q-Bench-Portrait: Benchmarking Multimodal Large Language Models on Portrait Image Quality Perception}

\author{
Sijing Wu$^{1,2}$,\; Yunhao Li$^{1,2}$,\; Zicheng Zhang$^{2}$,\; Qi Jia$^{2}$,\; Xinyue Li$^{1}$, \\
\vspace{0.3em}
Huiyu Duan$^{1}$,\; Xiongkuo Min$^{1}$,\; Guangtao Zhai$^{1,2}$
\affiliations
$^{1}$Shanghai Jiao Tong University
\affiliations
$^{2}$Shanghai Artificial Intelligence Laboratory
}

\begin{document}

\maketitle

\begin{abstract}

Recent advances in multimodal large language models (MLLMs) have demonstrated impressive performance on existing low-level vision benchmarks, which primarily focus on generic images. However, their capabilities to perceive and assess portrait images, a domain characterized by distinct structural and perceptual properties, remain largely underexplored.
To this end, we introduce \textbf{Q-Bench-Portrait}, the first holistic benchmark specifically designed for portrait image quality perception, comprising 2,765 image–question–answer triplets and featuring (1) diverse portrait image sources, including natural, synthetic distortion, AI-generated, artistic, and computer graphics images; (2) comprehensive quality dimensions, covering technical distortions, AIGC-specific distortions, and aesthetics; and (3) a range of question formats, including single-choice, multiple-choice, true/false, and open-ended questions, at both global and local levels.
Based on Q-Bench-Portrait, we evaluate 20 open-source and 5 closed-source MLLMs, revealing that although current models demonstrate some competence in portrait image perception, their performance remains limited and imprecise, with a clear gap relative to human judgments.
We hope that the proposed benchmark will foster further research into enhancing the portrait image perception capabilities of both general-purpose and domain-specific MLLMs.

\end{abstract}

\begin{figure*}
\centering
\includegraphics[width=\linewidth]{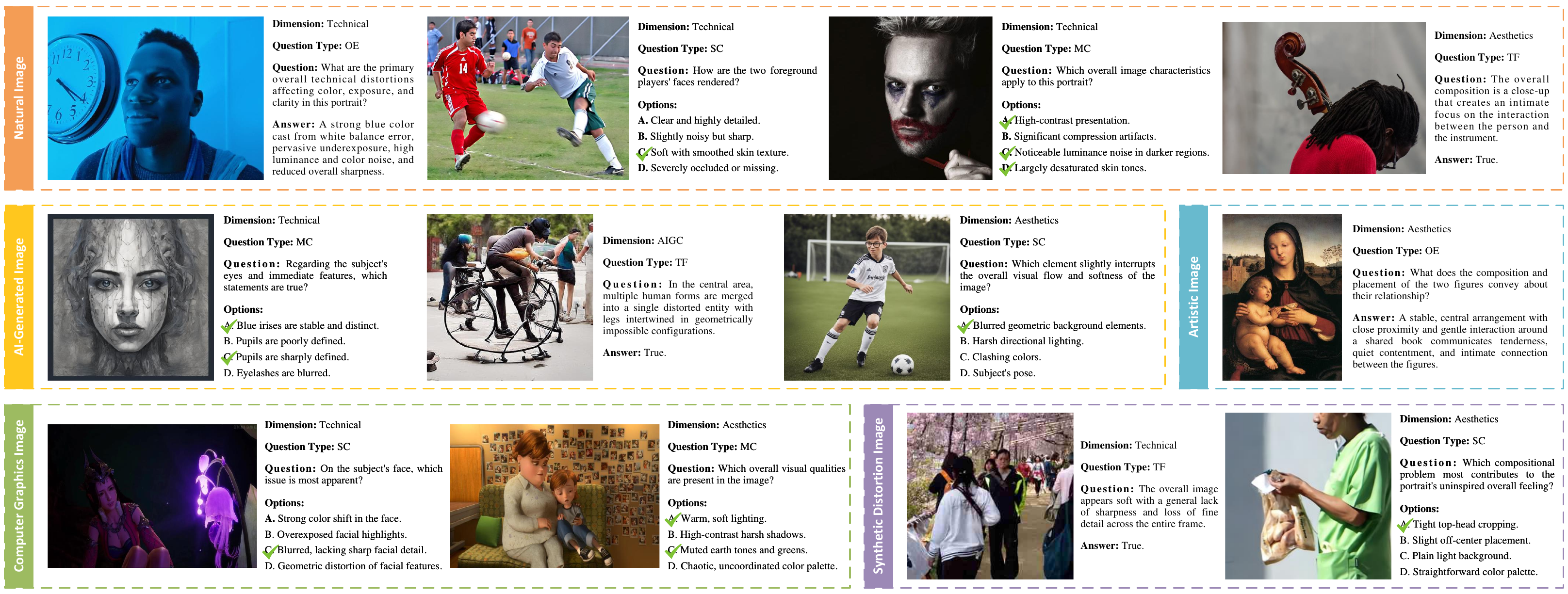}
\vspace{-5mm}
\caption{Examples from Q-Bench-Portrait. We present images and their corresponding question–answer pairs across different image categories. Notably, for each category, the images are associated with questions covering at least one of the three quality dimensions (\textit{i.e.}, technical distortions, AIGC-specific distortions, and aesthetics) and all four question types (\textit{i.e.}, SC, MC, TF, and OE).}
\label{fig:demo}
\end{figure*}

\input{tabs/bench_compare}

\input{tabs/source_dataset}

\section{Introduction}

Portrait images constitute a core visual domain centered on people rather than generic scenes. They are prevalent in real-world scenarios and underpin a wide range of applications such as photography and content creation. Unlike generic images, portrait images require fine-grained perception of subtle visual cues and nuanced human-related semantics, making their visual quality particularly perceptually sensitive \cite{liu2025human,wu2025hveval}.
With the rapid advancement of multimodal large language models (MLLMs) \cite{li2024llava,wang2025internvl3,Qwen3-VL,wu2024deepseek,qin2025ui,team2024gemini,singh2025openai}, their performance in high-level understanding and reasoning \cite{qin2025face,liu2025human}, as well as low-level quality perception on generic images \cite{wu2023q,zhang2025q}, has been consistently validated and shown to improve on existing benchmarks.
However, their capabilities in perceiving and assessing portrait image quality remain largely underexplored.

Existing benchmarks for MLLMs can be broadly categorized into low-level quality benchmarks and high-level semantic understanding benchmarks.
Low-level benchmarks, exemplified by Q-Bench \cite{wu2023q} and Q-Bench-Video \cite{zhang2025q}, evaluate perceptual quality assessment capabilities on generic images and videos. However, these benchmarks predominantly focus on generic content, which differs substantially from portrait images in both structural characteristics and human perceptual sensitivity.
In contrast, high-level benchmarks \cite{qin2025face,liu2025human} emphasize semantic understanding tasks such as reasoning, recognition, and multimodal comprehension, and have recently been extended to portrait-specific scenarios, underscoring the significance of treating human-centric visual content as a distinct domain for evaluation.
To date, no dedicated benchmark systematically focuses on the low-level quality perception of portrait images.

To address this gap, we present \textbf{Q-Bench-Portrait}, the first holistic benchmark specifically designed for portrait image quality perception (see Figure \ref{fig:demo} and Table \ref{tab:bench}).
Specifically, Q-Bench-Portrait consists of 2,765 image–question–answer triplets covering diverse portrait image sources, comprehensive quality dimensions, and various question types.
As illustrated in Table \ref{tab:source}, we collect portrait images from 17 diverse source datasets that cover five image types, including natural images (NIs), synthetic distortion images (SDIs), AI-generated images (AGIs), artistic images (AIs), and computer graphics images (CGIs). To maintain quality diversity, we uniformly sample images within each score range from the test sets.
Based on the source images, we construct questions along three quality dimensions to comprehensively characterize portrait image quality: technical distortions (\textit{e.g.}, noise and exposure), AI-generated content (AIGC)-specific distortions (\textit{e.g.}, anatomy and texture), and aesthetics (\textit{e.g.}, composition and lightning).
Moreover, we incorporate diverse question types, including single-choice (SC), multiple-choice (MC), true/false (TF), and open-ended (OE) questions, each formulated at both global and local scopes. To ensure benchmark quality, each question is reviewed and refined by human experts through multiple rounds of verification. As a result, we collect a total of 2,765 distinct portrait images, with each image corresponding to a single question.

Based on Q-Bench-Portrait, we systematically benchmark 20 open-source and 5 closed-source MLLMs on portrait image quality perception. 
Our benchmark-based analysis offers several insights into the current capabilities of MLLMs in portrait image quality perception.
First, open-source MLLMs, such as Qwen3-VL \cite{Qwen3-VL}, exhibit competitive and in some cases superior performance compared to closed-source models.
Second, fine-grained quality problems and complex question types remain challenging for current MLLMs.
Third, current MLLMs show limited capability in assessing the quality of non-natural portrait images, including AI-generated and computer graphics-based portraits.
Together, these observations underscore the need for dedicated evaluation of portrait image quality and motivate future research toward MLLMs with enhanced sensitivity to human-centric visual quality.

In summary, the main contributions of this paper are:
\begin{itemize}
    \item To the best of our knowledge, we are the first to benchmark MLLMs on the perceptual quality of portrait images, in contrast to existing benchmarks that primarily focus on generic images or high-level semantic understanding.

    \item We establish Q-Bench-Portrait, the first holistic benchmark specifically designed for portrait image perception, which comprises 2,765 Q\&A pairs covering diverse portrait image sources, comprehensive quality dimensions, and various question types and scopes.

    \item Based on Q-Bench-Portrait, we benchmark the portrait image perception capabilities of 25 MLLMs, revealing their strengths and limitations and providing insights into future research directions.
    
\end{itemize}

\section{Related Work}

\subsection{Multimodal Large Language Model}
The rapid development of multimodal large language models (MLLMs), such as LLaVA \cite{liu2023visual}, Gemini \cite{team2024gemini}, and ChatGPT \cite{singh2025openai}, has demonstrated remarkable capabilities in visual understanding and reasoning. By integrating vision and language, these models have been successfully applied to a variety of high-level computer vision tasks, including visual question answering, visual grounding, and complex visual reasoning. Beyond high-level semantic tasks, MLLMs are also expected to exhibit strong performance in low-level perception and understanding tasks, such as image and video quality assessment, highlighting the need for dedicated low-level benchmarks.

\subsection{Multimodal Benchmark}
To systematically evaluate the capabilities of MLLMs, a series of benchmarks have been proposed in recent years, which can be broadly categorized into high-level semantic understanding benchmarks and low-level quality benchmarks.
Conventional high-level benchmarks, such as MMBench \cite{liu2024mmbench}, primarily focus on evaluating semantic visual understanding across a diverse range of abilities. More recent benchmarks \cite{qin2025face,liu2025human} have extended this evaluation to portrait-specific scenarios, underscoring the importance of treating human-centric visual content as a distinct domain for assessment.
Meanwhile, low-level benchmarks \cite{wu2023q,zhang2025q} evaluate perceptual quality assessment capabilities on generic images and videos. However, these benchmarks primarily focus on generic images and largely overlook the evaluation of MLLMs’ quality perception capabilities on portrait images, which represent a significant and ubiquitous visual content domain. To address this gap, we introduce the Q-Bench-Portrait benchmark.

\subsection{Visual Quality Assessment}
Visual quality assessment aims to predict quality scores for various visual modalities, such as images, videos, and 3D content. In the early years, visual quality assessment methods \cite{mittal2012no} mainly rely on handcrafted features to extract quality-related information for quality prediction. With the development of deep learning, subsequent works utilize diverse convolutional neural network architectures \cite{su2020blindly} for quality regression. Subsequently, with the rise of Transformer-based architectures, numerous visual quality assessment methods \cite{yang2022maniqa} have emerged by modifying the Transformer network structure and attention mechanisms. In recent years, leveraging the powerful visual perception capabilities of MLLMs, MLLM-based quality assessment methods have demonstrated superior performance. Q-Align \cite{wu2023qalign} first applies MLLM to the image quality assessment task. Following this work, numerous studies have explored the use of MLLM for quality assessment across various domains, including user-generated content (UGC) videos \cite{ge2025lmm,zhang2025vq}, 3D content \cite{yang2025lmme3dhf,duan2025bmpcqa,li2025dhqa}, and human-centric images/videos \cite{wu2025fvq,li2025aghi,wu2025hveval}.

\begin{figure}[t]
\centering
\includegraphics[width=0.92\linewidth]{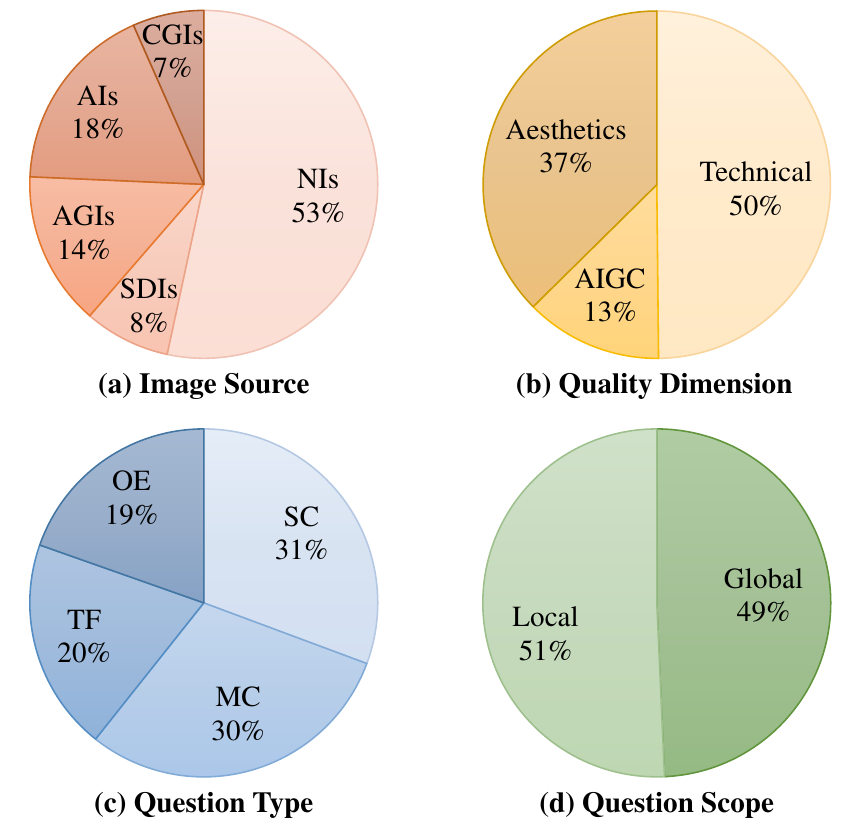}
\caption{Distributions of (a) image sources, (b) quality dimensions, (c) question types, and (d) question scopes.}
\label{fig:pie}
\end{figure}

\begin{figure}[t]
\centering
\includegraphics[width=\linewidth]{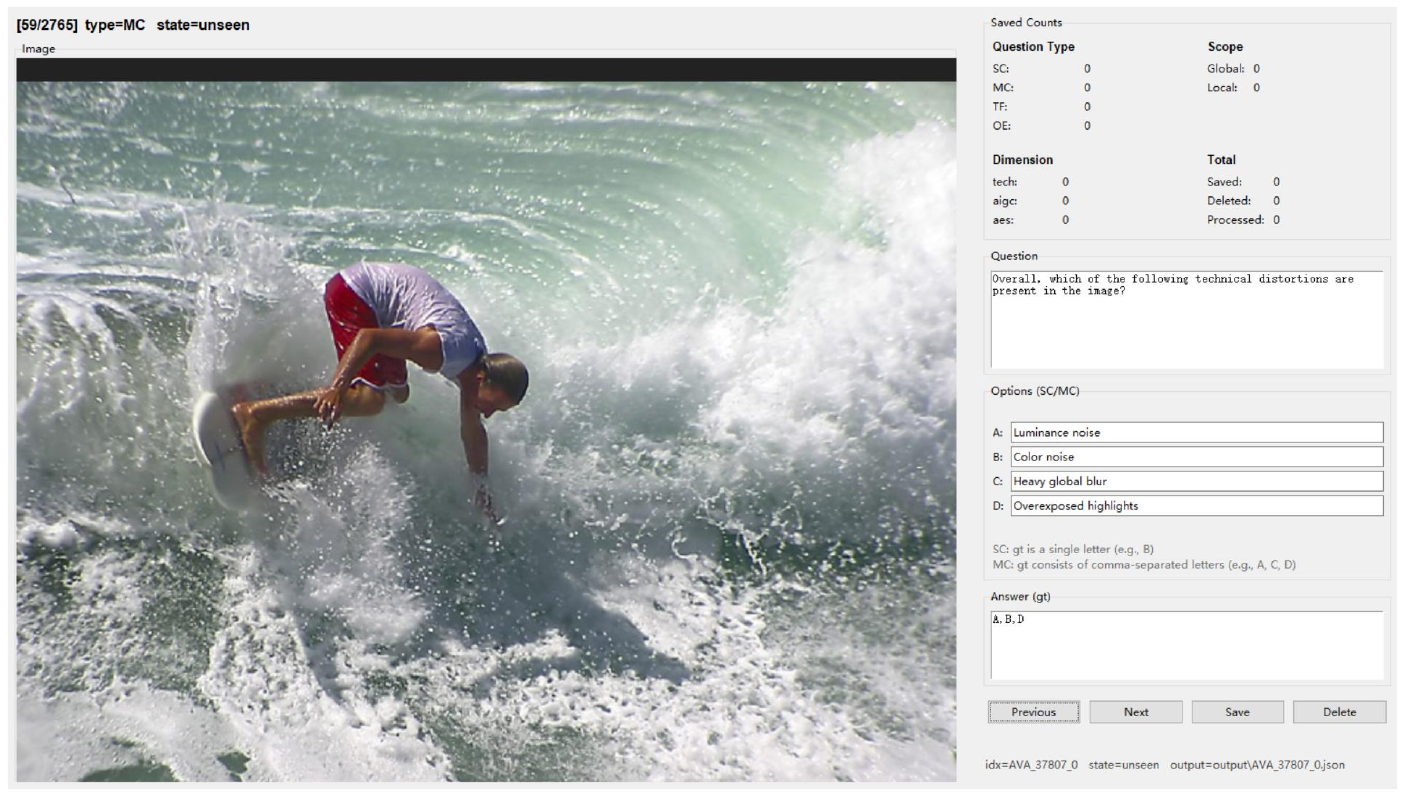}
\caption{Illustration of the GUI used to review and refine questions and answers in Q-Bench-Portrait.}
\label{fig:gui}
\end{figure}

\section{Q-Bench-Portrait}

In this section, we first describe the pipeline for collecting raw portrait images, then present the detailed quality dimensions and question types of the proposed benchmark, and finally explain the benchmark question construction and annotation pipeline.
An overview of the statistics of Q-Bench-Portrait is illustrated in Figure \ref{fig:pie}.

\input{tabs/tab_main}

\input{tabs/tab_sub_dim}

\input{tabs/tab_sub_imgtype_dim}

\input{tabs/tab_sub_scope}

\subsection{Portrait Image Collection}

As summarized in Table \ref{tab:source}, we initially collect source portrait images from image quality assessment and aesthetic quality assessment datasets. The main reasons are twofold: (1) these datasets contain diverse images spanning a wide range of quality scores; and (2) the provided quality/aesthetic scores in these datasets enable us to uniformly sample portrait images with diverse quality.

Specifically, we select 17 datasets as the raw image pool, including (1) natural image datasets: CGFIQA-40k \cite{chen2024dsl}, PIQ \cite{chahine2023image}, LiveBeauty \cite{li2025fpem}, MEBeauty \cite{lebedeva2022mebeauty}, KonIQ-10k \cite{hosu2020koniq}, SPAQ \cite{fang2020perceptual}, AVA \cite{murray2012ava}, TAD66K \cite{he2022rethinking}; (2) synthetic distortion image datasets: FIQA \cite{liu2024assessing}, PIPAL \cite{jinjin2020pipal}, KADID-10k \cite{lin2019kadid}; (3) AI-generated image datasets: AGHI-QA \cite{li2025aghi}, EvalMi-50k \cite{wang2025lmm4lmm}; (4) artistic image datasets: ArtEmis \cite{achlioptas2021artemis}, BAID \cite{yi2023towards}; and (5) computer graphics image datasets: CGIQA-6k \cite{zhang2023subjective}, NBU-CIQAD \cite{chen2021perceptual}. Among these datasets, KonIQ-10k, SPAQ, AVA, TAD66K, PIPAL, KADID-10k, EvalMi-50k, ArtEmis, BAID, CGIQA-6k, and NBU-CIQAD are general image datasets. 
To extract portrait images, we employ YOLOv9 to identify images containing visible persons, followed by MTCNN to detect visible portrait faces. After constructing the raw image pool, we uniformly sample portrait images from each dataset according to their perceptual quality or aesthetic scores.

\subsection{Quality Dimension}
The quality of portrait images can be influenced by multiple factors. We therefore define the following quality dimensions, which serve as guidelines for benchmark construction.

\noindent\textbf{Technical Distortions.} Technical distortions refer to low-level quality degradations in portrait images caused by imperfections in image acquisition, processing, compression, or generation. Typical technical distortions include blur, noise, compression artifacts, contrast degradation, exposure issues, color deviation, and related artifacts. Such distortions are commonly observed in natural images, synthetic distortion images, computer graphics images, and AI-generated images.

\noindent\textbf{AIGC-specific Distortions.} AIGC-specific distortions refer to unnatural image content and over-smoothed or inconsistent textures commonly observed in generated portrait images. Typical AIGC-specific distortions include geometric and structural errors, abnormal human body parts, texture artifacts, and regional inconsistency. Such distortions are unique to AI-generated images.

\noindent\textbf{Aesthetics.} Aesthetics refer to perceptual and affective issues related to the visual appeal of portrait images. Typical aesthetic aspects include visual composition, lighting, visual style, emotion, and user interest. This dimension is commonly involved in artistic images, natural images, AI-generated images, and computer graphics images.

\subsection{Question Design}

\noindent\textbf{Single-Choice Questions (SC).} This question type is widely used in existing MLLM benchmarks. Each question consists of a prompt posing a question and four answer options (phrases or sentences). This task requires MLLMs to select the most appropriate answer from the set of options that includes distractors, thereby comprehensively evaluating their ability to understand and distinguish diverse quality issues arising from different distortions.

\noindent\textbf{Multiple-Choice Questions (MC).} Considering that single-choice questions may not thoroughly evaluate the whole capabilities of MLLMs, we additionally construct multiple-choice questions, which are more challenging than single-choice ones. Each question also consists of a prompt posing a question and four answer options (phrases or sentences), and there may be more than one correct answer. By incorporating this type of question, we can evaluate whether current MLLMs are able to identify all relevant quality issues in portrait images. Moreover, these questions represent the most challenging component of our benchmark, encouraging future studies to address these challenges.

\noindent\textbf{True/False Questions (TF).} This question type requires MLLMs to make a binary judgment (True or False) based on a quality description sentence of a portrait image, aiming to evaluate their ability to assess portrait image quality. For this question type, we ensure that each description sentence contains relevant information corresponding to the given portrait image, thereby eliminating ambiguous or meaningless statements. In addition, we carefully control the ratio of True and False answers during the benchmark construction process to avoid severe class imbalance.

\noindent\textbf{Open-ended Questions (OE).} All the above question types focus on evaluating MLLMs' ability to select the correct answer options from predefined sets. However, in real-world scenarios, users may expect MLLMs to provide explainable quality descriptions of portrait images. To evaluate this type of ability, we introduce open-ended questions that do not restrict answers to predefined sets, such as ``Considering the whole image, how do global noise and compression influence the appearance of the subject's skin tones and the headband?" By incorporating this question type, we can more effectively assess MLLMs’ ability to perceive and articulate portrait image quality in practical applications.

\subsection{Benchmark Construction \& Quality Control}

Based on the curated portrait images, defined quality dimensions, and diverse question types, we construct Q-Bench-Portrait using a GPT-assisted, human-in-the-loop annotation strategy. Specifically, we first design detailed descriptions and category definitions for each of the three quality dimensions under the guidance of quality assessment experts. Then, we generate a caption for each portrait image under the specified quality dimension with the help of Gemini \cite{team2024gemini}. Based on the captions, we formulate four questions for each image by randomly selecting one question type, including two global-level questions and two local-level questions. Finally, all images and their associated questions and answers are manually reviewed and refined using the GUI illustrated in Figure \ref{fig:gui}. This human annotation process is conducted for at least two rounds to ensure data quality. Notably, a sufficiently large pool of candidate questions is constructed, allowing questions of excessively poor quality to be discarded during the review process.

\section{Experiments}
\subsection{Experimental Setting}

\noindent\textbf{Benchmarking MLLMs.} In our benchmark, 25 popular MLLMs are included for comprehensive evaluation, containing (1) 20 open-source MLLMs: InternVL series \cite{zhu2025internvl3,wang2025internvl3}, QwenVL series \cite{bai2025qwen2,Qwen3-VL}, DeepSeek-VL series \cite{lu2024deepseek,wu2024deepseek}, mPLUG-Owl3 \cite{ye2024mplug}, UI-TARS \cite{qin2025ui}, LLaVA-OneVision \cite{li2024llava}, Ministral-3 \cite{liu2026ministral}, Pixtral \cite{agrawal2024pixtral}; and (2) 5 closed-sourced MLLMs: Qwen3-VL-Plus \cite{Qwen3-VL}, Grok-4 \cite{grok4}, Claude-Sonnet-4.5 \cite{claude}, Gemini-2.5-Pro \cite{team2024gemini}, GPT-5.2 \cite{singh2025openai}.

\noindent\textbf{Evaluation Metrics.} For closed-ended questions, we manually define accuracy criteria for each question type. A question is considered correctly answered if MLLMs select all correct options for multiple-choice questions, select the correct option for single-choice questions, and output the correct judgment (True or False) for true/false questions. As for open-ended questions, we employ the latest ChatGPT \cite{singh2025openai} to compare the model-generated responses with the ground-truth answers and assign a discrete score ranging from 0 to 2. The prompt used for evaluating open-ended questions is provided in the supplementary material.

\begin{figure*}
\centering
\includegraphics[width=\linewidth]{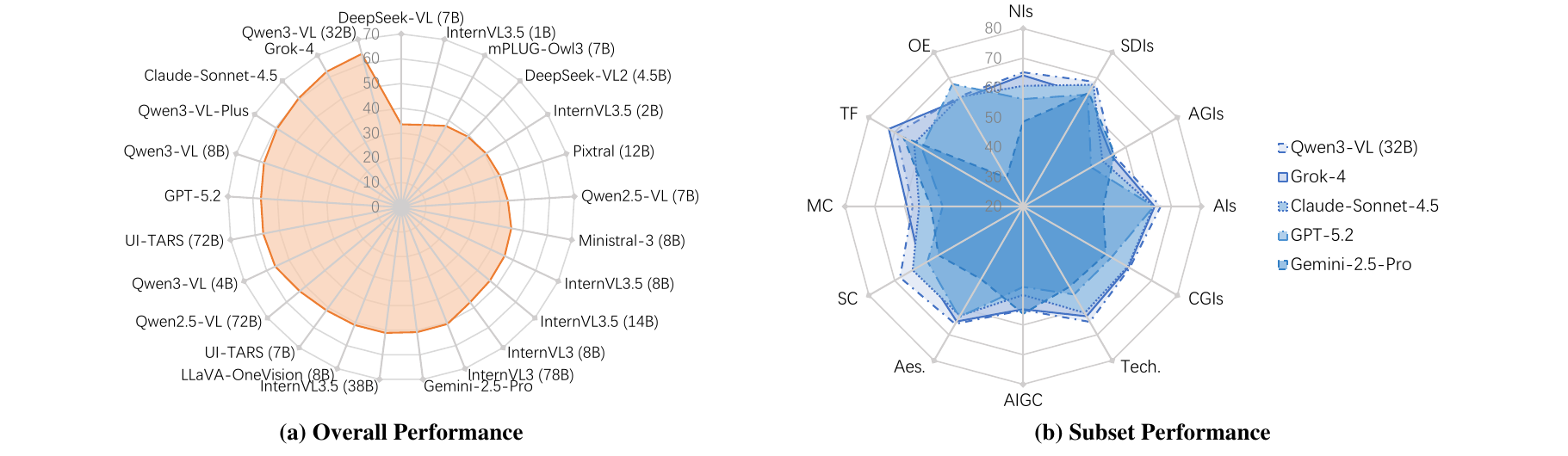}
\caption{Radar chart of benchmark results. (a) Comparison of overall performance across all MLLMs. (b) Comparison of five representative MLLMs across different subsets, including image types, quality dimensions, and question types.}
\label{fig:radar}
\end{figure*}

\subsection{Main Results} 

We summarize the overall quality perception performance and detailed subcategory performance on Q-Bench-Portrait in Table \ref{tab:main}. In addition, we report performance across various sub-dimensions in Table \ref{tab:sub_dim} (fine-grained quality dimensions), Table \ref{tab:sub_imgtype_dim} (quality dimensions across different image types), and Table \ref{tab:sub_scope} (global and local questions for each question type). Based on these results, our findings are summarized as follows.

\noindent\textbf{Overall performance: Qwen3-VL surpasses closed-source MLLMs.}
From the overall results reported in Table \ref{tab:main} and Figure \ref{fig:radar}, we can observe a surprising finding that Qwen3-VL-32B achieves the best overall performance, with an accuracy of 64.00\%, outperforming widely used closed-source MLLMs. Such results were not observed in Q-Bench \cite{wu2023q} in 2023, indicating that the performance gap between closed-source and open-source models has been steadily narrowing. 
Meanwhile, we can also observe that closed-source models still exhibit overall superior performance compared to the majority of open-source models. Among the closed-source models, Grok-4 achieves the highest score of 62.51\%, which is comparable to that of Qwen3-VL-32B.
Notably, even the best-performing MLLM attains only a relatively modest accuracy of 64.00\%, underscoring the substantial room for improvement in the portrait image quality perception capabilities of current MLLMs.

\noindent\textbf{Question type: Multiple-choice questions are more challenging for MLLMs.}
Based on the results in Table \ref{tab:main} and Figure \ref{fig:radar}, we can observe a consistent ranking of task difficulty in portrait image quality assessment for MLLMs. Multiple-choice questions pose the greatest challenge, followed by single-choice questions and open-ended questions, while true/false questions are the easiest. Notably, even the best-performing MLLM, Grok-4, achieves only 59.16\% accuracy on multiple-choice questions, indicating that current MLLMs still struggle with diverse and precise quality perception tasks.

\noindent\textbf{Image type: AI-generated and computer graphics images are more challenging for MLLMs.}
Table \ref{tab:main}, Table \ref{tab:sub_imgtype_dim}, and Figure \ref{fig:radar} present the quality perception performance of different image types, including natural images, synthetic distortion images, AI-generated images, artistic images, and computer graphics images.
Overall, most MLLMs exhibit relatively lower performance on AI-generated images and computer graphics images compared to other image domains. This performance gap is likely due to the fact that current MLLMs are predominantly trained on real-world images, which limits their ability to accurately perceive quality issues in less frequently encountered domains such as AI-generated and computer graphics images. These findings underscore the importance of placing greater emphasis on these image types in future research.

\noindent\textbf{Quality dimension: Blur and composition are challenging quality issues.}
As shown in Table \ref{tab:sub_dim}, performance across detailed quality sub-dimensions under technical distortions, AIGC-specific distortions, and aesthetics is reported.
For technical distortions, ``blur" exhibits relatively lower performance than other distortion types, with an accuracy of 62.24$\%$, while most MLLMs perform well in recognizing ``color" issues.
For AIGC-specific distortions, different MLLMs demonstrate varying performance on anatomy, texture, and semantics distortion problems. For aesthetics, most models perform worse on composition-related questions than on other sub-dimensions, indicating that current models struggle to understand the spatial arrangement and structural organization of visual elements in portrait images.

\noindent\textbf{Local vs. Global: Local questions are more challenging than global questions.}
Local and global questions constitute two fundamental question scopes in quality benchmarks \cite{wu2023q,zhang2025q}. Global questions evaluate the ability of perceiving quality issues at the level of the entire portrait image, whereas local questions measure the ability of perceiving quality issues within specific regions of a portrait image.
As shown in Table~\ref{tab:sub_scope}, we can observe that most MLLMs exhibit lower performance on local questions than on global questions, highlighting their limited capability in understanding localized portrait image quality.

\section{Conclusion}

In this paper, we introduce \textbf{Q-Bench-Portrait}, the first holistic benchmark specifically designed for portrait image quality perception, comprising 2,765 image–question–answer triplets and featuring (1) diverse portrait image sources, including natural, synthetic distortion, AI-generated, artistic, and computer graphics images; (2) comprehensive quality dimensions, covering technical distortions, AIGC-specific distortions, and aesthetics; and (3) a range of question formats, including single-choice, multiple-choice, true/false, and open-ended questions, at both global and local levels. Based on Q-Bench-Portrait, we evaluate 25 open-source and closed-source MLLMs, revealing that although current models demonstrate some competence in portrait image perception, their performance remains limited and imprecise. We hope that Q-Bench-Portrait will facilitate future research on advancing portrait image perception in both general-purpose and domain-specific MLLMs.






\bibliographystyle{named}
\bibliography{ijcai26}

\end{document}

%% file: tabs/bench_compare.tex
\begin{table*}[t]
\centering
\renewcommand\arraystretch{1.1}

\resizebox{\linewidth}{!}{
\begin{tabular}{lccccccc}
\toprule

Benchmark	&	Domain	&	Modality	&	Level	&	Source Data	&	QA Pairs	&	Data Type	&	Question Type	\\
\midrule

Q-Bench \cite{wu2023q}	&	General	&	Image	&	Low-level	&	2,990	&	2,990	&	NIs, SDIs, AGIs	&	SC, TF, OE	\\
AesBench \cite{huang2024aesbench}	&	General	&	Image	&	High-level	&	2,800	&	8,400	&	NIs, AGIs, AIs	&	SC, TF, OE	\\
Q-Bench-Video \cite{zhang2025q}	&	General	&	Video	&	Mixed	&	1,800	&	2,378	&	NVs, AGVs, CGVs	&	SC, TF, OE	\\
MedQ-Bench \cite{liu2025medq}	&	Medical	&	Image	&	Low-level	&	-	&	3,308	&	NIs, SDIs, AGIs	&	SC, TF, OE	\\
\hdashline

Face-Human-Bench \cite{qin2025face}	&	Human	&	Image	&	High-level	&	-	&	3,600	&	NIs	&	SC	\\

\rowcolor[gray]{.92}
\textbf{Q-Bench-Portrait (Ours)}	&	Human	&	Image	&	Mixed	&	2,765	&	2,765	&	NIs, SDIs, AGIs, AIs, CGIs	&	SC, MC, TF, OE	\\

\bottomrule
\end{tabular}
}
\vspace{-1mm}
\caption{Summary and comparison of related benchmarks. The abbreviations ``NVs", ``AGVs", ``CGVs" represent natural videos, AI-generated videos, and computer graphics videos, respectively.}
\vspace{-2mm}
\centering
\label{tab:bench}
\end{table*}

%% file: tabs/source_dataset.tex
\begin{table}[t]
\centering
\renewcommand\arraystretch{1.1}

\resizebox{\linewidth}{!}{
\begin{tabular}{lccccc}
\toprule

Dataset	&	Annotation	&	Full Size	&	Sampled Size \\
\midrule

$\diamondsuit$ CGFIQA-40k	&	Perceptual Score	&	40,000	&	149	\\
$\diamondsuit$ PIQ	&	Perceptual Score	&	5,116	&	184	\\
$\diamondsuit$ LiveBeauty	&	Attractiveness Score	&	10,000	&	196	\\
$\diamondsuit$ MEBeauty	&	Attractiveness Score	&	2,550	&	100	\\
$\diamondsuit$ KonIQ-10k\dag	&	Perceptual Score	&	10,073	&	147	\\
$\diamondsuit$ SPAQ\dag	&	Perceptual Score	&	11,125	&	118	\\
$\diamondsuit$ AVA\dag	&	Aesthetic Score	&	255,508	&	500	\\
$\diamondsuit$ TAD66K\dag	&	Aesthetic Score	&	66,000	&	82	\\
\hdashline

$\spadesuit$ FIQA	&	Perceptual Score	&	42,125	&	145	\\
$\spadesuit$ PIPAL\dag	&	Perceptual Score	&	29,000	&	44	\\
$\spadesuit$ KADID-10k\dag	&	Perceptual Score	&	10,125	&	32	\\
\hdashline

$\heartsuit$ AGHI-QA	&	Perceptual Score	&	4,000	&	217	\\
$\heartsuit$ EvalMi-50K\dag	&	Perceptual Score	&	50,000	&	179	\\
\hdashline

$\clubsuit$ ArtEmis\dag	&	Description	&	81,446	&	327	\\
$\clubsuit$ BAID\dag	&	Aesthetic Score	&	60,337	&	162	\\
\hdashline

\raisebox{0.4ex}{\scalebox{0.6}{$\bigcirc$}} CGIQA-6k\dag	&	Perceptual Score	&	6,000	&	166	\\
\raisebox{0.4ex}{\scalebox{0.6}{$\bigcirc$}} NBU-CIQAD\dag	&	Perceptual Score	&	1,800	&	17	\\

\bottomrule
\end{tabular}
}
\centering
\caption{Overview of the diverse portrait image source datasets in Q-Bench-Portrait. $\diamondsuit$, $\spadesuit$, $\heartsuit$, $\clubsuit$, and \raisebox{0.4ex}{\scalebox{0.6}{$\bigcirc$}} denote NIs, SDIs, AGIs, AIs, and CGIs, respectively. \dag indicates datasets that contain generic images, from which only portrait images are sampled.}
\label{tab:source}
\end{table}

%% file: tabs/tab_main.tex
\begin{table*}[t]
\centering
\renewcommand\arraystretch{1.05}
\small
\resizebox{\textwidth}{!}{
\begin{tabular}{l ccccc ccc cccc c}
\toprule
\multirow{2}{*}{\textbf{Model}} & \multicolumn{5}{c}{\textbf{Image Sources}} & \multicolumn{3}{c}{\textbf{Quality Dimensions}} & \multicolumn{4}{c}{\textbf{Question Types}} & \multirow{2}{*}{\textbf{Overall}\,$\uparrow$} \\
\cmidrule(lr){2-6} \cmidrule(lr){7-9} \cmidrule(lr){10-13}
& NIs & SDIs & AGIs & AIs & CGIs & Tech. & AIGC & Aes. & SC & MC & TF & OE \\
\midrule

\rowcolor[gray]{.92}
\multicolumn{14}{l}{\textit{Baseline}} \\
\hdashline

Random guess (\textit{w/o} MC and OE)	&	34.76 	&	34.62 	&	35.42 	&	34.89 	&	33.23 	&	34.84 	&	35.71 	&	34.32 	&	25.00 	&	-	&	50.00 	&	-	&	34.78	\\

\hdashline
\rowcolor[gray]{.92}
\multicolumn{14}{l}{\textit{Open-source MLLMs}} \\
\hdashline

InternVL3.5 (1B) \cite{wang2025internvl3}	&	36.01 	&	28.05 	&	26.39 	&	40.90 	&	28.96 	&	32.69 	&	26.63 	&	39.31 	&	40.57 	&	16.87 	&	51.19 	&	34.69 	&	34.39 	\\
InternVL3.5 (2B) \cite{wang2025internvl3}	&	42.38 	&	33.71 	&	31.82 	&	47.96 	&	33.88 	&	40.20 	&	31.87 	&	44.10 	&	47.05 	&	26.63 	&	55.60 	&	36.81 	&	40.60 	\\
Qwen3-VL (4B) \cite{Qwen3-VL}	&	57.22 	&	57.69 	&	45.45 	&	60.84 	&	56.56 	&	55.81 	&	44.62 	&	60.59 	&	57.67 	&	52.41 	&	61.65 	&	54.06 	&	56.17 	\\
DeepSeek-VL2 (4.5B) \cite{wu2024deepseek}	&	41.50 	&	36.20 	&	26.52 	&	46.42 	&	32.51 	&	37.55 	&	24.79 	&	46.32 	&	29.60 	&	31.33 	&	60.92 	&	44.46 	&	39.20 	\\
Qwen2.5-VL (7B) \cite{bai2025qwen2}	&	45.33 	&	41.86 	&	32.45 	&	50.41 	&	30.87 	&	42.53 	&	32.29 	&	47.68 	&	55.54 	&	19.88 	&	56.51 	&	45.94 	&	43.15 	\\
DeepSeek-VL (7B) \cite{lu2024deepseek}	&	33.57 	&	31.22 	&	33.71 	&	36.20 	&	28.69 	&	31.57 	&	34.70 	&	35.78 	&	39.98 	&	0.36 	&	65.87 	&	41.79 	&	33.54 	\\
mPLUG-Owl3 (7B) \cite{ye2024mplug}	&	38.41 	&	33.26 	&	31.82 	&	45.09 	&	27.60 	&	34.58 	&	31.59 	&	43.47 	&	28.77 	&	31.33 	&	63.49 	&	34.59 	&	37.52 	\\
UI-TARS (7B) \cite{qin2025ui} &	48.44 	&	50.90 	&	54.55 	&	59.10 	&	48.09 	&	48.26 	&	54.25 	&	54.55 	&	58.73 	&	39.88 	&	66.42 	&	42.34 	&	51.37 	\\
LLaVA-OneVision (8B) \cite{li2024llava}	&	51.29 	&	49.32 	&	43.94 	&	60.12 	&	43.17 	&	48.29 	&	42.63 	&	57.74 	&	62.15 	&	31.69 	&	66.06 	&	48.52 	&	51.10 	\\
Qwen3-VL (8B) \cite{Qwen3-VL}	&	60.40 	&	63.57 	&	43.56 	&	63.19 	&	52.73 	&	59.51 	&	42.92 	&	61.75 	&	57.90 	&	52.65 	&	68.99 	&	56.46 	&	58.23 	\\
InternVL3 (8B) \cite{zhu2025internvl3}	 &	47.90 	&	47.74 	&	36.74 	&	56.24 	&	41.53 	&	45.72 	&	34.99 	&	53.72 	&	53.18 	&	38.43 	&	58.35 	&	40.77 	&	47.34 	\\
InternVL3.5 (8B) \cite{wang2025internvl3}	&	47.15 	&	45.70 	&	33.46 	&	55.21 	&	41.26 	&	44.99 	&	33.57 	&	51.89 	&	52.71 	&	31.08 	&	59.45 	&	45.39 	&	46.11 	\\
Ministral-3 (8B) \cite{liu2026ministral}	&	47.43 	&	39.14 	&	34.47 	&	51.23 	&	42.35 	&	45.32 	&	33.85 	&	49.03 	&	51.42 	&	24.58 	&	62.20 	&	50.18 	&	45.24 	\\
Pixtral (12B) \cite{agrawal2024pixtral}	&	43.46 	&	39.59 	&	32.32 	&	47.24 	&	36.61 	&	42.45 	&	31.73 	&	44.29 	&	42.57 	&	25.66 	&	60.18 	&	46.68 	&	41.77 	\\
InternVL3.5 (14B) \cite{wang2025internvl3}	&	45.93 	&	45.02 	&	39.27 	&	55.62 	&	43.44 	&	43.72 	&	38.10 	&	52.95 	&	50.59 	&	36.75 	&	58.53 	&	42.71 	&	46.46 	\\
Qwen3-VL (32B) \cite{Qwen3-VL}	&	\textbf{65.24} 	& \textbf{68.55} 	&	\underline{55.05} 	&	\textbf{66.36} 	&	\underline{61.48} 	&	\textbf{64.99} 	&	\underline{54.96}	&	\textbf{65.76} 	&	\textbf{67.92} 	&	57.23 	&	\underline{69.36} 	& \underline{62.82}  &	\textbf{64.00} 	\\
InternVL3.5 (38B) \cite{wang2025internvl3}	&	51.66 	&	49.77 	&	40.66 	&	60.84 	&	44.26 	&	49.27 	&	38.95 	&	57.59 	&	55.66 	&	43.01 	&	61.65 	&	45.57 	&	51.07 	\\
Qwen2.5-VL (72B) \cite{bai2025qwen2} &	54.03 	&	55.88 	&	48.86 	&	56.54 	&	42.62 	&	52.58 	&	50.14 	&	54.88 	&	57.43 	&	43.98 	&	64.59 	&	48.89 	&	53.13 	\\
UI-TARS (72B) \cite{qin2025ui}	&	53.73 	&	54.75 	&	59.47 	&	61.76 	&	\textbf{62.84} 	&	53.77 	&	\textbf{59.49} 	&	59.53 	&	57.43 	&	55.54 	&	68.26 	&	45.48 	&	56.65 	\\
InternVL3 (78B)	\cite{zhu2025internvl3} &	51.93 	&	51.36 	&	41.04 	&	56.34 	&	45.36 	&	49.85 	&	40.37 	&	55.27 	&	51.42 	&	44.82 	&	64.22 	&	44.83 	&	50.67 	\\

\hdashline
\rowcolor[gray]{.92}
\multicolumn{14}{l}{\textit{Closed-source MLLMs}} \\
\hdashline

Qwen3-VL-Plus \cite{Qwen3-VL}	&	60.70 	&	64.48 	&	53.91 	&	59.51 	&	53.83 	&	60.52 	&	53.97 	&	59.67 	&	57.31 	&	\underline{58.43} 	&	64.04 	&	59.32 	&	59.37 	\\
Grok-4 \cite{grok4}	&	\underline{64.19} 	&	63.35 	&	54.29 	&	\underline{64.31} 	&	60.93 	&	\underline{62.81} 	&	54.67 	&	\underline{64.80} 	&	59.91 	&	\textbf{59.16} 	&	\textbf{72.29} 	&	61.90 	&	\underline{62.51} 	\\
Claude-Sonnet-4.5 \cite{claude}	&	60.70 	&	\underline{67.42} 	&	50.88 	&	64.01 	&	60.66 	&	61.50 	&	49.86 	&	62.57 	&	\underline{62.97}	&	55.06 	&	62.39 	&	62.64 	&	60.42 	\\
Gemini-2.5-Pro \cite{team2024gemini}	&	48.61 	&	64.25 	&	\textbf{55.56} 	&	47.03 	&	52.19 	&	51.27 	&	56.23 	&	48.36 	&	52.95 	&	51.08 	&	66.61 	&	31.18 	&	50.81 	\\
GPT-5.2 \cite{singh2025openai}	&	56.23 	&	63.57 	&	46.46 	&	64.01 	&	54.37 	&	54.28 	&	47.17 	&	63.10 	&	57.19 	&	47.11 	&	59.45 	&	\textbf{67.71} 	&	56.67 	\\

\bottomrule
\end{tabular}
}
\caption{Experimental results on Q-Bench-Portrait for evaluating the portrait image quality perception capabilities of MLLMs. The abbreviations ``Tech.", ``AIGC", and ``Aes." denote technical distortions, AIGC-specific distortions, and aesthetics, respectively. The best and runner-up performances are bold and underlined, respectively.}
\vspace{-2mm}
\label{tab:main}
\end{table*}

%% file: tabs/tab_sub_dim.tex
\begin{table*}[t]
\centering
\renewcommand\arraystretch{1.05}
\small
\resizebox{\textwidth}{!}{
\begin{tabular}{l ccccc ccc cccc}
\toprule
\multirow{2}{*}{\textbf{Model}} & \multicolumn{5}{c}{\textbf{Technical Distortions}} & \multicolumn{3}{c}{\textbf{AIGC-specific Distortions}} & \multicolumn{4}{c}{\textbf{Aesthetics}} \\
\cmidrule(lr){2-6} \cmidrule(lr){7-9} \cmidrule(lr){10-13}
& Noise & Blur & Exposure & Color & Artifacts & Anatomy & Texture & Semantics & Emotion & Composition & Lighting & Style \\
\midrule

\rowcolor[gray]{.92}
\multicolumn{13}{l}{\textit{Baseline}} \\
\hdashline

Random guess (\textit{w/o} MC and OE)	&	37.80 	&	33.54 	&	36.52 	&	36.05 	&	35.76 	&	36.49 	&	35.10 	&	36.11 	&	35.57 	&	32.95 	&	35.60 	&	34.33 	\\

\hdashline
\rowcolor[gray]{.92}
\multicolumn{13}{l}{\textit{Open-source MLLMs}} \\
\hdashline

InternVL3.5 (1B) \cite{wang2025internvl3}	&	37.71 	&	30.99 	&	35.51 	&	39.63 	&	31.22 	&	29.24 	&	25.98 	&	20.97 	&	40.35 	&	36.98 	&	44.79 	&	37.58 	\\
InternVL3.5 (2B) \cite{wang2025internvl3}	&	39.83 	&	38.82 	&	45.17 	&	46.95 	&	38.83 	&	27.54 	&	32.84 	&	41.94 	&	44.21 	&	43.37 	&	47.24 	&	42.48 	\\
Qwen3-VL (4B) \cite{Qwen3-VL}	&	53.81 	&	54.29 	&	56.53 	&	66.46 	&	58.12 	&	38.98 	&	47.55 	&	46.77 	&	65.27 	&	58.35 	&	61.66 	&	55.88 	\\
DeepSeek-VL2 (4.5B) \cite{wu2024deepseek}	&	38.14 	&	34.97 	&	42.33 	&	39.02 	&	42.89 	&	23.31 	&	25.25 	&	27.42 	&	49.68 	&	43.12 	&	48.47 	&	45.75 	\\
Qwen2.5-VL (7B) \cite{bai2025qwen2}	&	54.24 	&	40.12 	&	46.88 	&	42.68 	&	41.37 	&	30.08 	&	33.82 	&	30.65 	&	50.00 	&	45.95 	&	50.61 	&	44.44 	\\
DeepSeek-VL (7B) \cite{lu2024deepseek}	&	45.34 	&	29.13 	&	37.50 	&	40.24 	&	24.37 	&	35.59 	&	33.33 	&	40.32 	&	39.55 	&	33.91 	&	38.34 	&	30.39 	\\
mPLUG-Owl3 (7B) \cite{ye2024mplug}	&	29.66 	&	32.05 	&	36.36 	&	39.63 	&	44.16 	&	33.05 	&	31.37 	&	27.42 	&	47.43 	&	38.82 	&	48.77 	&	42.16 	\\
UI-TARS (7B) \cite{qin2025ui}	&	52.97 	&	49.44 	&	53.41 	&	51.22 	&	34.77 	&	59.32 	&	52.21 	&	48.39 	&	58.04 	&	50.12 	&	56.13 	&	57.52 	\\
LLaVA-OneVision (8B) \cite{li2024llava} 	&	47.03 	&	44.72 	&	54.83 	&	59.76 	&	53.05 	&	47.88 	&	37.99 	&	53.23 	&	61.90 	&	56.14 	&	59.82 	&	51.31 	\\
Qwen3-VL (8B) \cite{Qwen3-VL}	&	59.32 	&	57.33 	&	63.35 	&	68.90 	&	61.17 	&	39.83 	&	41.42 	&	\textbf{64.52} 	&	64.79 	&	58.48 	&	65.34 	&	60.46 	\\
InternVL3 (8B) \cite{zhu2025internvl3}	&	48.73 	&	45.09 	&	41.76 	&	53.05 	&	46.95 	&	33.47 	&	36.03 	&	33.87 	&	58.52 	&	52.09 	&	50.00 	&	52.29 	\\
InternVL3.5 (8B) \cite{wang2025internvl3}	&	44.92 	&	43.11 	&	49.72 	&	51.83 	&	45.69 	&	34.75 	&	30.64 	&	48.39 	&	52.73 	&	49.88 	&	56.75 	&	50.33 	\\
Ministral-3 (8B) \cite{liu2026ministral} &	45.34 	&	44.53 	&	46.88 	&	54.88 	&	43.15 	&	30.08 	&	35.29 	&	38.71 	&	50.64 	&	48.03 	&	49.08 	&	48.37 	\\
Pixtral (12B) \cite{agrawal2024pixtral}	&	44.92 	&	38.57 	&	50.57 	&	56.71 	&	43.65 	&	33.47 	&	31.62 	&	25.81 	&	47.75 	&	40.91 	&	43.87 	&	46.73 	\\
InternVL3.5 (14B) \cite{wang2025internvl3}	&	44.49 	&	40.31 	&	46.02 	&	54.27 	&	50.76 	&	30.93 	&	40.44 	&	50.00 	&	58.20 	&	46.56 	&	57.36 	&	54.58 	\\
Qwen3-VL (32B) \cite{Qwen3-VL}	&	\textbf{69.92} 	&	\textbf{62.24} 	&	\textbf{70.45} 	&	\textbf{74.39} 	&	64.47 	&	56.78 	&	53.68 	&	56.45 	&	\textbf{69.29} 	&	\underline{62.78} &	\textbf{67.18} 	&	\underline{65.03} 	\\
InternVL3.5 (38B) \cite{wang2025internvl3}	&	45.76 	&	48.14 	&	44.60 	&	64.63 	&	53.81 	&	45.76 	&	33.82 	&	46.77 	&	61.58 	&	52.09 	&	60.43 	&	61.11 	\\
Qwen2.5-VL (72B) \cite{bai2025qwen2}	&	58.90 	&	47.08 	&	62.50 	&	62.80 	&	58.12 	&	50.85 	&	50.49 	&	45.16 	&	56.27 	&	53.32 	&	57.98 	&	52.94 	\\
UI-TARS (72B) \cite{qin2025ui}	&	50.00 	&	54.97 	&	59.66 	&	61.59 	&	42.64 	&	\textbf{63.98} 	&	\textbf{58.33} 	&	50.00 	&	61.74 	&	56.02 	&	62.58 	&	61.11 	\\
InternVL3 (78B) \cite{zhu2025internvl3}	&	49.15 	&	47.08 	&	51.70 	&	62.80 	&	54.57 	&	36.86 	&	41.67 	&	45.16 	&	60.13 	&	52.70 	&	57.06 	&	50.33 	\\

\hdashline
\rowcolor[gray]{.92}
\multicolumn{13}{l}{\textit{Closed-source MLLMs}} \\
\hdashline

Qwen3-VL-Plus \cite{Qwen3-VL}	&	58.90 	&	58.63 	&	60.51 	&	70.73 	&	\underline{64.97} 	&	51.27 	&	\underline{56.37} 	&	48.39 	&	61.41 	&	58.97 	&	57.98 	&	59.80 	\\
Grok-4 \cite{grok4}	&	62.29 	&	\underline{60.25} 	&	63.64 	&	\underline{71.34}	&	\textbf{69.29} 	&	57.20 	&	52.70 	& \underline{58.06} &	\underline{68.49} 	&	62.65 	&	65.64 	& 62.09 \\
Claude-Sonnet-4.5 \cite{claude}	&	\underline{69.07} 	&	59.38 	&	\underline{65.34} 	&	67.68 	&	59.64 	&	47.46 	&	51.23 	&	50.00 	&	62.06 	&	60.69 	&	\underline{66.26} 	&	\textbf{64.71} 	\\
Gemini-2.5-Pro \cite{team2024gemini}	&	55.51 	&	54.47 	&	46.31 	&	45.73 	&	42.39 	&	\underline{59.75} 	&	54.41 	&	54.84 	&	49.20 	&	45.70 	&	47.24 	&	54.90 	\\
GPT-5.2 \cite{singh2025openai}	&	58.90 	&	52.92 	&	52.27 	&	62.80 	&	55.33 	&	44.07 	&	47.79 	&	54.84 	&	61.74 	&	\textbf{64.37} 	&	61.04 	&	\textbf{64.71} 	\\

\bottomrule
\end{tabular}
}
\caption{Experimental results on Q-Bench-Portrait across fine-grained quality dimensions. The best and runner-up performances are bold and underlined, respectively.}
\label{tab:sub_dim}
\vspace{-3mm}
\end{table*}

%% file: tabs/tab_sub_imgtype_dim.tex
\begin{table*}[t]
\centering
\small
\resizebox{\textwidth}{!}{
\begin{tabular}{l ccc ccc cccc c ccc}
\toprule
\multirow{2}{*}{\textbf{Model}} & \multicolumn{3}{c}{\textbf{NIs}} & \multicolumn{3}{c}{\textbf{SDIs}} & \multicolumn{4}{c}{\textbf{AGIs}} & \multicolumn{1}{c}{\textbf{AIs}} & \multicolumn{3}{c}{\textbf{CGIs}} \\
\cmidrule(lr){2-4} \cmidrule(lr){5-7} \cmidrule(lr){8-11} \cmidrule(lr){12-12} \cmidrule(lr){13-15}
& Tech. & Aes. & Overall & Tech. & Aes. & Overall & Tech. & AIGC & Aes. & Overall & Aes. & Tech. & Aes. & Overall \\
\midrule

\rowcolor[gray]{.92}
\multicolumn{15}{l}{\textit{Baseline}} \\
\hdashline

Random guess	&	35.05 	&	33.81 	&	34.76 	&	34.91 	&	34.32 	&	34.62 	&	31.82 	&	35.71 	&	33.33 	&	35.42 	&	34.89 	&	33.56 	&	30.56 	&	33.23 	 	\\

\hdashline
\rowcolor[gray]{.92}
\multicolumn{15}{l}{\textit{Open-source MLLMs}} \\
\hdashline

InternVL3.5 (1B) \cite{wang2025internvl3}	&	34.21 	&	40.97 	&	36.01 	&	28.51 	&	27.57 	&	28.05 	&	18.00 	&	26.63 	&	33.33 	&	26.39 	&	40.90 	&	27.56 	&	37.04 	&	28.96 		\\
InternVL3.5 (2B) \cite{wang2025internvl3}	&	41.27 	&	45.42 	&	42.38 	&	40.79 	&	26.17 	&	33.71 	&	36.00 	&	31.87 	&	25.00 	&	31.82 	&	47.96 	&	33.01 	&	38.89 	&	33.88 		\\
Qwen3-VL (4B) \cite{Qwen3-VL}	&	56.65 	&	58.78 	&	57.22 	&	53.07 	&	62.62 	&	57.69 	&	46.00 	&	44.62 	&	61.11 	&	45.45 	&	60.84 	&	53.53 	&	74.07 	&	56.56 		\\
DeepSeek-VL2 (4.5B) \cite{wu2024deepseek}	&	39.52 	&	46.95 	&	41.50 	&	32.89 	&	39.72 	&	36.20 	&	34.00 	&	24.79 	&	50.00 	&	26.52 	&	46.42 	&	27.88 	&	59.26 	&	32.51 	 	\\
Qwen2.5-VL (7B) \cite{bai2025qwen2}	&	44.88 	&	46.56 	&	45.33 	&	42.98 	&	40.65 	&	41.86 	&	34.00 	&	32.29 	&	33.33 	&	32.45 	&	50.41 	&	27.24 	&	51.85 	&	30.87 	 	\\
DeepSeek-VL (7B) \cite{lu2024deepseek}	&	32.64 	&	36.13 	&	33.57 	&	28.95 	&	33.64 	&	31.22 	&	18.00 	&	34.70 	&	36.11 	&	33.71 	&	36.20 	&	28.21 	&	31.48 	&	28.69 		\\
mPLUG-Owl3 (7B) \cite{ye2024mplug}	&	36.47 	&	43.77 	&	38.41 	&	28.95 	&	37.85 	&	33.26 	&	26.00 	&	31.59 	&	44.44 	&	31.82 	&	45.09 	&	26.92 	&	31.48 	&	27.60 		\\
UI-TARS (7B) \cite{qin2025ui}	&	47.97 	&	49.75 	&	48.44 	&	50.00 	&	51.87 	&	50.90 	&	56.00 	&	54.25 	&	58.33 	&	54.55 	&	59.10 	&	47.76 	&	50.00 	&	48.09 		\\
LLaVA-OneVision (8B) \cite{li2024llava}	&	49.63 	&	55.85 	&	51.29 	&	45.18 	&	53.74 	&	49.32 	&	54.00 	&	42.63 	&	55.56 	&	43.94 	&	60.12 	&	40.38 	&	59.26 	&	43.17 		\\
Qwen3-VL (8B) \cite{Qwen3-VL}	&	60.85 	&	59.16 	&	60.40 	&	62.28 	&	\underline{64.95} 	&	63.57 	&	48.00 	&	42.92 	&	50.00 	&	43.56 	&	63.19 	&	50.00 	&	68.52 	&	52.73 	\\
InternVL3 (8B) \cite{zhu2025internvl3}	&	46.40 	&	52.04 	&	47.90 	&	48.68 	&	46.73 	&	47.74 	&	58.00 	&	34.99 	&	41.67 	&	36.74 	&	56.24 	&	36.86 	&	68.52 	&	41.53 	\\
InternVL3.5 (8B) \cite{wang2025internvl3}	&	46.17 	&	49.87 	&	47.15 	&	44.30 	&	47.20 	&	45.70 	&	34.00 	&	33.57 	&	30.56 	&	33.46 	&	55.21 	&	39.10 	&	53.70 	&	41.26 	\\
Ministral-3 (8B)	&	46.86 	&	48.98 	&	47.43 	&	39.91 	&	38.32 	&	39.14 	&	34.00 	&	33.85 	&	47.22 	&	34.47 	&	51.23 	&	40.38 	&	53.70 	&	42.35 	\\
Pixtral (12B) \cite{agrawal2024pixtral}	&	43.77 	&	42.62 	&	43.46 	&	40.79 	&	38.32 	&	39.59 	&	34.00 	&	31.73 	&	41.67 	&	32.32 	&	47.24 	&	35.90 	&	40.74 	&	36.61 	\\
InternVL3.5 (14B) \cite{wang2025internvl3}	&	44.28 	&	50.51 	&	45.93 	&	44.74 	&	45.33 	&	45.02 	&	40.00 	&	38.10 	&	61.11 	&	39.27 	&	55.62 	&	39.74 	&	64.81 	&	43.44 \\
Qwen3-VL (32B) \cite{Qwen3-VL}	&	\textbf{65.60} 	&	\underline{64.25} 	&	\textbf{65.24} 	&	\underline{71.05} 	&	\textbf{65.89} 	&	\textbf{68.55} 	&	52.00 	&	54.96 	&	\underline{61.11} 	&	\underline{55.05} 	&	\textbf{66.36} 	&	58.33 	&	\textbf{79.63} 	&	\underline{61.48} \\
InternVL3.5 (38B) \cite{wang2025internvl3}	&	50.55 	&	54.71 	&	51.66 	&	50.44 	&	49.07 	&	49.77 	&	52.00 	&	38.95 	&	58.33 	&	40.66 	&	60.84 	&	39.10 	&	\underline{74.07} 	&	44.26 	\\
Qwen2.5-VL (72B) \cite{bai2025qwen2}	&	53.74 	&	54.83 	&	54.03 	&	62.28 	&	49.07 	&	55.88 	&	38.00 	&	50.14 	&	38.89 	&	48.86 	&	56.54 	&	39.74 	&	59.26 	&	42.62 \\
UI-TARS (72B) \cite{qin2025ui}	&	52.95 	&	55.85 	&	53.73 	&	53.07 	&	56.54 	&	54.75 	&	46.00 	&	\textbf{59.49} 	&	77.78 	&	59.47 	&	61.76 	&	61.22 	&	72.22 	&	\textbf{62.84} 	\\
InternVL3 (78B) \cite{zhu2025internvl3}	&	51.20 	&	53.94 	&	51.93 	&	50.44 	&	52.34 	&	51.36 	&	36.00 	&	40.37 	&	61.11 	&	41.04 	&	56.34 	&	42.31 	&	62.96 	&	45.36 \\

\hdashline
\rowcolor[gray]{.92}
\multicolumn{15}{l}{\textit{Closed-source MLLMs}} \\
\hdashline

Qwen3-VL-Plus \cite{Qwen3-VL}	&	60.94 	&	60.05 	&	60.70 	&	68.42 	&	60.28 	& 64.48 &	\underline{56.00} 	&	53.97 	&	50.00 	&	53.91 	&	59.51 	&	52.56 	&	61.11 	&	53.83 		\\
Grok-4 \cite{grok4}	&	\underline{63.02} 	&	\textbf{67.43} 	&	\underline{64.19} 	&	70.18 	&	56.07 	&	63.35 	&	40.00 	&	54.67 	&	\textbf{66.67} 	&	54.29 	&	\underline{64.31} 	&	\underline{59.62} 	&	68.52 	&	60.93 	\\
Claude-Sonnet-4.5 \cite{claude}	&	60.53 	&	61.20 	&	60.70 	&	70.61 	&	64.02 	&	\underline{67.42} 	&	\textbf{64.00} 	&	49.86 	&	52.78 	&	50.88 	&	64.01 	&	\textbf{61.22} 	&	57.41 	&	60.66 	\\
Gemini-2.5-Pro \cite{team2024gemini}	&	49.49 	&	46.18 	&	48.61 	&	\textbf{71.93} 	&	56.07 	&	64.25 	&	42.00 	&	\underline{56.23} 	&	61.11 	&	\textbf{55.56} 	&	47.03 	&	50.00 	&	64.81 	&	52.19 \\
GPT-5.2 \cite{singh2025openai}	&	53.88 	&	62.72 	&	56.23 	&	66.23 	&	60.75 	&	63.57 	&	36.00 	&	47.17 	&	47.22 	&	46.46 	&	64.01 	&	51.28 	&	72.22 	&	54.37 	\\

\bottomrule
\end{tabular}
}
\caption{Experimental results on Q-Bench-Portrait across different quality dimensions under various image sources. The best and runner-up performances are bold and underlined, respectively.}
\label{tab:sub_imgtype_dim}
\vspace{-1mm}
\end{table*}

%% file: tabs/tab_sub_scope.tex
\begin{table}[t]
\centering
\renewcommand\arraystretch{1.05}
\small
\resizebox{\linewidth}{!}{
\begin{tabular}{l cc cc cc cc}
\toprule
\multirow{2}{*}{\textbf{Model}} & \multicolumn{2}{c}{\textbf{SC}} & \multicolumn{2}{c}{\textbf{MC}} & \multicolumn{2}{c}{\textbf{TF}} & \multicolumn{2}{c}{\textbf{OE}}\\
\cmidrule(lr){2-3} \cmidrule(lr){4-5} \cmidrule(lr){6-7} \cmidrule(lr){8-9}
& Global & Local & Global & Local & Global & Local & Global & Local \\
\midrule

\rowcolor[gray]{.92}
\multicolumn{9}{l}{\textit{Baseline}} \\
\hdashline

Random guess	&	25.00 	&	25.00 	&	- 	&	- 	&	50.00 	&	50.00 	&	- 	&	- 	\\

\hdashline
\rowcolor[gray]{.92}
\multicolumn{9}{l}{\textit{Open-source MLLMs}} \\
\hdashline

InternVL3.5 (1B)	&	41.04 	&	40.05 	&	19.72 	&	13.78 	&	51.01 	&	51.34 	&	37.70 	&	32.21 	\\
InternVL3.5 (2B)	&	49.43 	&	44.47 	&	27.61 	&	25.56 	&	54.25 	&	56.71 	&	42.01 	&	32.55 	\\
Qwen3-VL (4B)	&	60.54 	&	54.55 	&	55.45 	&	49.12 	&	63.16 	&	60.40 	&	58.20 	&	50.67 	\\
DeepSeek-VL2 (4.5B)	&	28.34 	&	30.96 	&	31.55 	&	31.08 	&	63.97 	&	58.39 	&	47.75 	&	41.78 	\\
Qwen2.5-VL (7B)	&	58.28 	&	52.58 	&	18.33 	&	21.55 	&	60.73 	&	53.02 	&	48.57 	&	43.79 	\\
DeepSeek-VL (7B)	&	46.71 	&	32.68 	&	0.70 	&	0.00 	&	65.18 	&	66.44 	&	46.52 	&	37.92 	\\
mPLUG-Owl3 (7B)	&	28.34 	&	29.24 	&	34.34 	&	28.07 	&	62.75 	&	64.09 	&	38.73 	&	31.21 	\\
UI-TARS (7B)	&	60.32 	&	57.00 	&	39.44 	&	40.35 	&	65.59 	&	67.11 	&	45.49 	&	39.77 	\\
LLaVA-OneVision (8B)	&	63.27 	&	60.93 	&	30.86 	&	32.58 	&	66.40 	&	65.77 	&	49.80 	&	47.48 	\\
Qwen3-VL (8B)	&	61.90 	&	53.56 	&	56.15 	&	48.87 	&	69.23 	&	68.79 	&	60.25 	&	53.36 	\\
InternVL3 (8B)	&	55.78 	&	50.37 	&	39.21 	&	37.59 	&	57.89 	&	58.72 	&	44.26 	&	37.92 	\\
InternVL3.5 (8B)	&	53.97 	&	51.35 	&	35.27 	&	26.57 	&	59.92 	&	59.06 	&	47.13 	&	43.96 	\\
Ministral-3 (8B)	&	56.69 	&	45.70 	&	23.20 	&	26.07 	&	62.75 	&	61.74 	&	54.10 	&	46.98 	\\
Pixtral (12B)	&	46.94 	&	37.84 	&	25.52 	&	25.81 	&	68.83 	&	53.02 	&	51.43 	&	42.79 	\\
InternVL3.5 (14B)	&	50.34 	&	50.86 	&	35.27 	&	38.35 	&	56.68 	&	60.07 	&	48.57 	&	37.92 	\\
Qwen3-VL (32B)	&	\textbf{71.88} 	&	\textbf{63.64} 	&	\textbf{61.48} 	&	52.63 	&	\underline{73.28} 	&	66.11 	&	\underline{66.60} 	&	59.73 	\\
InternVL3.5 (38B)	&	55.33 	&	56.02 	&	42.00 	&	44.11 	&	57.09 	&	65.44 	&	48.16 	&	43.46 	\\
Qwen2.5-VL (72B)	&	60.09 	&	54.55 	&	46.64 	&	41.10 	&	67.21 	&	62.42 	&	51.02 	&	47.15 	\\
UI-TARS (72B)	&	59.86 	&	54.79 	&	53.60 	&	\underline{57.64} 	&	68.02 	&	68.46 	&	44.88 	&	45.97 	\\
InternVL3 (78B)	&	49.43 	&	53.56 	&	44.55 	&	45.11 	&	67.21 	&	61.74 	&	47.13 	&	42.95 	\\

\hdashline
\rowcolor[gray]{.92}
\multicolumn{9}{l}{\textit{Closed-source MLLMs}} \\
\hdashline

Qwen3-VL-Plus	&	60.32 	&	54.05 	&	57.08 	&	\textbf{59.90} 	&	68.83 	&	60.07 	&	61.68 	&	57.38 	\\
Grok-4	&	60.09 	&	59.71 	&	\underline{61.02} 	&	57.14 	&	\textbf{76.52} 	&	\underline{68.79} 	&	66.19 	&	58.39 	\\
Claude-Sonnet-4.5	&	\underline{63.49} 	&	\underline{62.41}	&	59.63 	&	50.13 	&	63.97 	&	61.07 	&	63.73 	&	\underline{61.74} 	\\
Gemini-2.5-Pro	&	51.93 	&	54.05 	&	54.99 	&	46.87 	&	62.35 	&	\textbf{70.13} 	&	27.66 	&	34.06 	\\
GPT-5.2	&	57.60 	&	56.76 	&	48.26 	&	45.86 	&	63.16 	&	56.38 	&	\textbf{70.08} 	&	\textbf{65.77} 	\\

\bottomrule
\end{tabular}
}
\caption{Experimental results on Q-Bench-Portrait across global and local question focuses for different question types. The best and runner-up performances are bold and underlined, respectively.}
\label{tab:sub_scope}
\vspace{-1mm}
\end{table}